\documentclass{llncs}
\usepackage{graphicx}
\usepackage{todonotes}
\usepackage{comment}
\usepackage{enumitem}
\usepackage{microtype}
\usepackage{setspace}
\usepackage[hyphens]{url}
\usepackage{hyperref}
\title{Neural Networks for Keyword Spotting on IoT Devices}
\author{Rakesh Dhakshinamurthy}
\institute{Dublin City University \\
\email{rakesh.dhakshinamurthy2@mail.dcu.ie}}

\begin{document}
\maketitle
\begin{abstract}

We explore Neural Networks (NNs) for keyword spotting (KWS) on IoT devices like smart speakers and wearables. Since we target to execute our NN on a constrained memory and computation footprint, we propose a CNN design that. (i) uses a limited number of multiplies. (ii) uses a limited number of model parameters.
\end{abstract}

\section{Introduction} \label{Intro}

Nowadays the speech-related NNs are becoming exponentially popular since it is an essential feature in a broad-spectrum of IoT devices. In IoT devices like wearables and smart speakers, true hands-free speech recognition is achieved by continuously listening to spot specific keywords. After spotting the specified keywords, the local or online voice understanding model is woken up, and the user's voice is streamed to the model. This keyword spotting (KWS) Models run on all the devices that are equipped with digital voice assistants like Alexa. Even in the advanced voice-based systems and smart speaker prototypes, such KWS models are integrated. For example, Sensory wake word engine was used in the recent study \cite{aics19smartspeaker}. Here, a face recognition algorithm was trained using Deep Neural Network and deployed on their modern Alexa smart speaker prototype. This model, without disturbing the smart speaker routine can detect and identify a human face, then start the Alexa voice service only when an authorized face is present in the live video frames. Similarly, another recent study \cite{aivision} also used the Sensory wake word engine to spot the Alexa wake word. Here, a DL and Open CV based object detection model was deployed in their smart speaker. After waking up the device by calling out the Alexa command, whenever the user calls out \emph{Alexa, ask Friday what she sees}, the smart speaker camera turns on, then executes the deployed model and calls out the names of detected objects as a response to the user command. The study \cite{smarthearingaid} presents a microphone array and voice algorithm-based smart hearing aid. We recommend the authors to add a KWS spotting mechanism to the software deployed in their prototype to improve the operating time of their device.

Since the above tiny devices and others have limited memory footprint and low computational power, the current KWS systems need tuning to comfortably execute on constrained setups and without disturbing the routine of the devices. To efficiently deploy large models on such devices, we recommend advanced model optimization methods such as \emph{RC-NN} \cite{rcenniot2020} and \emph{Edge2Train} \cite{edge2trainiot2020}. In the recent wearable-based study named \emph{Avoid touching your face} \cite{covidaway2020}, the authors have optimized their hand-to-face motion-detecting CNNs using RCE-NN to enable the resource-friendly execution of the use-case model on a wide range of smartwatches. Similarly, we recommend optimizing the adaptive strategy model \cite{wfiot2020} to make it comfortably execute within the limited resource of mobile IoT devices and carry out the model intended task of improving the wireless communication quality of the target devices. Also, the parts of this optimization pipeline are applicable to unsupervised methods \cite{unsupervised}.

In this paper, we take motivation from \cite{arik2017convolutional} to design CNNs for KWS use case. We propose a CNN based approach since CNNs have shown better performance than DNNs and also has a smaller model size. In our proposed design, we first limit the overall computation of our KWS model. i.e., we limit the model parameters and multiplies. Here, we first implement a CNN architecture that does not perform pooling but rather performs filter striding in the frequency domain. Second, we limit the total parameters during the design of the KWS model.

\section{Keyword Spotting on IoT Devices}

\begin{figure}[t]
\centering
 \includegraphics[width=15cm,height=7cm,keepaspectratio]{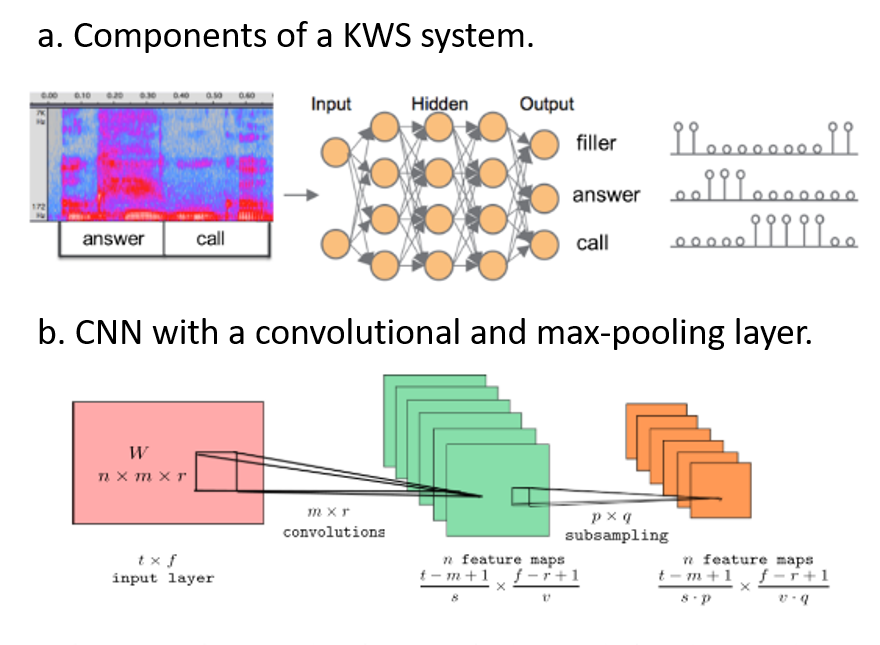}
\caption{a. Shows the components of a KWS system (left to right) (i) Feature Extraction (ii) Deep Neural Network (iii) Posterior Handling. b. Shows the architecture of a typical CNN with a convolutional and max-pooling layer.}
\label{fig1}  
\end{figure}

A block diagram of the KWS system \cite{chen2014small} used in this work is shown in Fig. \ref{fig1}. The first feature extraction component \emph{computes 40-dimensional log-mel filterbank features every 25ms with a 10ms frame shift}. Next, in each frame, we stack 25 frames to the left and 10 frames to the right and provide this as an input to DNN which is the second component. This DNN architecture has 3 hidden layers with 128 hidden units plus a softmax layer, where each hidden layer uses a ReLU nonlinearity. The softmax is the output layer containing one target output for each of the words that need to be detected from the audio signal stream. This model's weights are trained to optimize a cross-entropy criterion using the distributed asynchronous gradient descent. The final component is the posterior handling module. Here for each frame, posterior scores from the DNN are combined into a single score.

\section{Keyword Spotting CNN Design}

Here we present a CNN design that is our proposed alternative to the DNN described in the previous Section.

\subsection{Typical CNN Architecture}

\begin{figure}[t]
\centering
 \includegraphics[width=9cm,height=7cm,keepaspectratio]{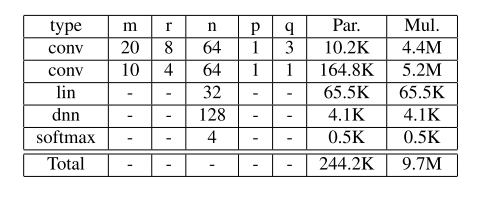}
\caption{Traditional CNN Architecture used for the KWS task. Par. is the number of parameters and Mul. is the number of multipliers used by each part of the CNN.}
\label{fig2}  
\end{figure}

A typical convolutional architecture shown in Fig. \ref{fig1} has been heavily used on many challenges \cite{sainath2013deep}. When the log-mel input ($t × f = 32 × 40$) is fed into the model, the first layer has a filter size with a specified frequency (freq $r = 9$). The common method is to choose a filter size in time spanning two-thirds of the overall input size. Multiplication is performed by striding this filter by $s = 1$ and $v = 1$ across both frequency and time domains. Next, non-overlapping max-pooling is performed with a pooling region of $q = 3$. The second filter has a filter of $r = 4$ in frequency, and no max-pooling is performed. Since we target limited resource devices, we need to keep the number of parameters as low as possible. We show the architecture of a traditional CNN in Fig. \ref{fig2}. Here, as shown, the model has 2 convolutional, one linear low-rank, and one DNN layer. Currently, this architecture has a huge number of multiplies in the convolutional layers due to the 3-D inputs spanning across all the three frequency, time, and feature maps. Hence, not feasible for power-constrained small-footprint devices where multiplies are limited. In the below sections, we present alternative CNN architectures to address the tasks of limiting parameters or multiplies.

\subsection{Limiting Multiplies}

\begin{figure}[t]
\centering
 \includegraphics[width=9cm,height=7cm,keepaspectratio]{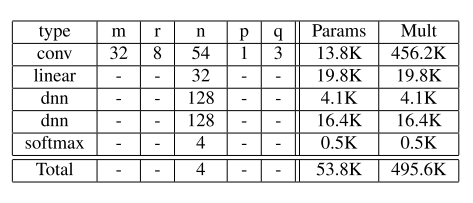}
\caption{Optimized CNN Architecture that performs the same KWS task but has a lesser number of Parameters and Multipliers than original CNN.}
\label{fig3}  
\end{figure}

To obtain a resource-friendly CNN we need to limit the number of multiplies. One solution to limit multiplies is to have a single convolutional layer, and also have the time filter span all of the time. The output of this convolutional layer should be passed to a linear low-rank layer followed by two DNN layers. Fig. \ref{fig3} shows an architecture with only one convolutional layer. Here, the number of multiplies is cut by a factor of 10, compared to the CNN from Fig. \ref{fig2}.

\subsection{Limiting Model Parameters}

Increasing model parameters leads to performance improvements at the cost of model size. In this section, we explore CNN architectures by limiting the model size to 250K but we do not limit the multiplies like in the previous section. We can improve performance by increasing the input features. Keeping parameters fixed by doing so requires exploring sampling of input audio signal in time and frequency. When designing models for acoustic use cases, the called words that we want to classify occur over within a very short time-duration, making pooling reduce the model detection accuracy. But the keywords like Alexa occur over a longer time-span (50-100ms). Thus, we propose to improve the resource friendliness of models by sub-sampling the signal in time using striding or pooling. In the future, we shall explore the conventional sub-sampling in time with longer acoustic units.

\section{Conclusion}

In this paper, we presented a CNN design that is best suited for the KWS use case. When comparing CNNs with DNNs, when we limit the number of multiplies and model parameters of CNNs, we hypothesize to achieve performance improvements in both clean and noisy environments. In the future, we shall evaluate our proposed design and open-source the trained model.

% --- BIBLIOGRAPHY
\begin{spacing}{0.92}
\bibliography{aics-sample.bib}

\begin{thebibliography}{10}
\providecommand{\url}[1]{\texttt{#1}}
\providecommand{\urlprefix}{URL }

\bibitem{arik2017convolutional}
Arik, S.O., Kliegl, M., Child, R., Hestness, J., Gibiansky, A., Fougner, C.,
  Prenger, R., Coates, A.: Convolutional recurrent neural networks for
  small-footprint keyword spotting. arXiv preprint arXiv:1703.05390  (2017)

\bibitem{chen2014small}
Chen, G., Parada, C., Heigold, G.: Small-footprint keyword spotting using deep
  neural networks. In: 2014 IEEE International Conference on Acoustics, Speech
  and Signal Processing (ICASSP). pp. 4087--4091. IEEE (2014)

\bibitem{sainath2013deep}
Sainath, T.N., Mohamed, A.r., Kingsbury, B., Ramabhadran, B.: Deep
  convolutional neural networks for lvcsr. In: 2013 IEEE international
  conference on acoustics, speech and signal processing. pp. 8614--8618. IEEE
  (2013)

\bibitem{wfiot2020}
Sudharsan, B., Breslin, J.G., Ali, M.I.: Adaptive strategy to improve the
  quality of communication for iot edge devices. In: 6th {IEEE} World Forum on
  Internet of Things, WF-IoT 2020, New Orleans, LA, USA, June 2-16, 2020. pp.
  1--6. {IEEE} (2020), \url{https://doi.org/10.1109/WF-IoT48130.2020.9221276}

\bibitem{edge2trainiot2020}
Sudharsan, B., Breslin, J.G., Ali, M.I.: Edge2train: a framework to train
  machine learning models (svms) on resource-constrained iot edge devices. In:
  Davidsson, P., Langheinrich, M. (eds.) IoT '20: 10th International Conference
  on the Internet of Things, Malm{\"{o}}, Sweden, October 6-9, 2020. pp.
  6:1--6:8. {ACM} (2020), \url{https://doi.org/10.1145/3410992.3411014}

\bibitem{rcenniot2020}
Sudharsan, B., Breslin, J.G., Ali, M.I.: {RCE-NN:} a five-stage pipeline to
  execute neural networks (cnns) on resource-constrained iot edge devices. In:
  Davidsson, P., Langheinrich, M. (eds.) IoT '20: 10th International Conference
  on the Internet of Things, Malm{\"{o}}, Sweden, October 6-9, 2020. pp.
  5:1--5:8. {ACM} (2020), \url{https://doi.org/10.1145/3410992.3411005}

\bibitem{smarthearingaid}
Sudharsan, B., Chockalingam, M.: A microphone array and voice algorithm based
  smart hearing aid. CoRR  abs/1908.07324 (2019),
  \url{http://arxiv.org/abs/1908.07324}

\bibitem{aics19smartspeaker}
Sudharsan, B., Corcoran, P., Ali, M.I.: Smart speaker design and implementation
  with biometric authentication and advanced voice interaction capability. In:
  Curry, E., Keane, M.T., Ojo, A., Salwala, D. (eds.) Proceedings for the 27th
  {AIAI} Irish Conference on Artificial Intelligence and Cognitive Science,
  Galway, Ireland, December 5-6, 2019. {CEUR} Workshop Proceedings, vol. 2563,
  pp. 305--316. CEUR-WS.org (2019),
  \url{http://ceur-ws.org/Vol-2563/aics\_29.pdf}

\bibitem{aivision}
Sudharsan, B., Kumar, S.P., Dhakshinamurthy, R.: Ai vision: Smart speaker
  design and implementation with object detection custom skill and advanced
  voice interaction capability. In: 2019 11th International Conference on
  Advanced Computing (ICoAC). pp. 97--102. IEEE (2019)

\bibitem{covidaway2020}
Sudharsan, B., Sundaram, D., Breslin, J.G., Ali, M.I.: Avoid touching your
  face: {A} hand-to-face 3d motion dataset (covid-away) and trained models for
  smartwatches. In: Davidsson, P., Langheinrich, M., Linde, P., Mayer, S.,
  Casado{-}Mansilla, D., Spikol, D., Kraemer, F.A., Russo, N.L. (eds.) IoT '20
  Companion: 10th International Conference on the Internet of Things Companion,
  Malm{\"{o}} Sweden, October 6-9, 2020. pp. 7:1--7:9. {ACM} (2020),
  \url{https://doi.org/10.1145/3423423.3423433}

\bibitem{unsupervised}
{Verma}, P., {Sudharsan}, B., {Chakravarthi}, B.R., {O'Riordan}, C., {Hill},
  S.: Unsupervised method to analyze playing styles of epl teams using ball
  possession-position data. In: 2020 6th International Conference on Advanced
  Computing and Communication Systems (ICACCS). pp. 58--64 (2020)

\end{thebibliography}
\bibliographystyle{splncs03}
\end{spacing}

\end{document}